\documentclass[11pt]{article}
\usepackage[letterpaper,margin=1in]{geometry}
\usepackage[T1]{fontenc}
\usepackage{lmodern}
\usepackage{microtype}
\usepackage{amsmath,amssymb}
\usepackage{graphicx}
\usepackage{booktabs}
\usepackage[table]{xcolor}
\usepackage{caption}
\usepackage{authblk}
\usepackage[numbers,sort&compress]{natbib}
\usepackage{setspace}
\usepackage[hidelinks]{hyperref}
\hypersetup{pdftitle={Accuracy is not compatibility: two Evo 2 checkpoints hinge on one FP8 matrix multiplication},
  pdfauthor={Xun Ding and Ziyuan Qin}}
\usepackage{url}
\newcommand{\code}[1]{\texttt{#1}}

\title{Accuracy is not compatibility: two Evo~2 checkpoints hinge on one FP8 matrix multiplication}
\author[1]{Xun Ding\thanks{ORCID 0000-0001-7719-7268. Correspondence:
\href{mailto:dxun1@alumni.jh.edu}{dxun1@alumni.jh.edu}}}
\author[2]{Ziyuan Qin\thanks{\href{mailto:zqin30@emory.edu}{zqin30@emory.edu}}}
\affil[1]{Independent researcher}
\affil[2]{Emory University, United States of America}
\date{}

\begin{document}
\maketitle

\begin{abstract}
FP8 formats specify how values are stored, not how products are summed; this gap can break
checkpoints. Evo~2, a DNA language model, documents a Hopper FP8 requirement for its 1B and 40B
checkpoints. On two Blackwell (sm\_120) GPUs both fail the developers' forward test (the 7B passes); on an
RTX~5070~Ti the 40B's next-base accuracy is 25.7\%, near guessing. Greater accuracy breaks them, as BioNeMo
saw for the 40B on Hopper: natively, 99.996\% of one projection's outputs are correctly rounded (H100: about
30\%), and fp32 accumulation of the same operands fails. Our H100 FP8 GEMM emulator, mmaemu (bit-exact
on captured 1B projections), localises the dependence to the 1B's first projection (of 21), where exact sums
fail but pass when shrunk by the H100's 1.4\% (measured in-sample), not 2.1\%. The 40B fails with only its
first projection native. Emulating only that projection, weights unchanged, raises 40B \textit{BRCA1}
variant-classification AUROC from 0.434 to 0.869 (H100: 0.868) on the RTX~5070~Ti, despite a narrow forward-test
miss, and 0.481 to 0.867 on an RTX~PRO~6000. We tested one model family but not training origin;
releases should specify reference arithmetic.
\end{abstract}

\section{Introduction}

When a released low-precision checkpoint runs on a new GPU or kernel, which arithmetic properties must the new
implementation preserve? The number format does not say: the OCP FP8 specification defines how values are
encoded and excludes arithmetic on them from its scope \cite{ofp8_2023}. Tensor cores differ in how many bits
they keep while accumulating FP8 products, 13 fraction bits on measured Hopper and Ada GPUs against 25 on the
B200 and RTX~PRO~6000 \cite{khattak2026accurate}, and libraries choose how partial sums are split and combined
\cite{nvidia_cublas_12_8}. That extra intermediate precision can change a program's results is well known
\cite{monniaux2008}, and changing the GPU version can change the outputs of large language models
\cite{yuan2025nondeterminism}. More accurate FP8 accumulation is a sound choice in training
\cite{deepseekai2024v3}; we ask whether a checkpoint with fixed weights keeps its
function when its accumulation changes, even towards greater accuracy.

Evo~2 is a substantial contribution to long-context genomic modelling \cite{brixi2026evo2}. The Arc Institute and its collaborators share weights, inference software, tests
with explicit tolerances and evaluation notebooks \cite{evo2readme2026}, and variant effects can be scored
zero-shot from changes in sequence likelihood \cite{benegas2023gpn,nguyen2024evo,brixi2026evo2}. The inference
code uses FP8 input projections in the Hyena blocks of StripedHyena~2
\cite{vortex,ku2025stripedhyena2,nvidia_transformer_engine}. The README lists FP8 on a Hopper GPU as a
requirement for the 1B, 20B and 40B checkpoints and advises validating outputs after hardware changes
\cite{evo2readme2026}; our experiments explore portability beyond that documented requirement.

Public reports describe failures of these upstream tests on consumer Blackwell GPUs and B200s, alongside a passing 7B
\cite{evo2issue157,evo2issue132}. NVIDIA's BioNeMo documentation notes that enabling more accurate accumulation breaks the original
40B even on Hopper and reports the original 1B working on
a B300 under its own stack and evaluation \cite{bionemo_evo2_docs}. A fine-tuning project attributes 20B/40B
failures to Hopper accumulation and localises a dependence to block~23 \cite{l1joseph_evo2ft}. The phenomenon
is therefore known; missing is a controlled path from an arithmetic change to a localised dependence and back
to recovery with the weights fixed.

We call the upstream Transformer Engine (TE) FP8 path \emph{official}, and the H100 PCIe and RTX~4060 running
it our \emph{reference GPUs}; on one measured projection their outputs lean towards zero, consistent with
truncating accumulation. mmaemu \cite{ding2026mmaemu}, our emulator of their FP8 matrix
multiplications (GEMMs), lets us change the arithmetic of chosen projections while the executing GPU, weights
and FP8 operands stay fixed: in effect, bisecting a model by arithmetic. Unlike instruction-level models such as
MMA-Sim \cite{xie2025mmasim}, mmaemu emulates whole library GEMM calls. Five findings follow: the more accurate GEMM is the incompatible one; one early
projection carries most of the dependence; in the 1B, that projection passes with a slight shrinkage like the
reference's, not with exact sums or more accumulator bits; downstream of it, the native difference collapsed the
1B but matched-size random perturbations did not; and restoring that projection alone recovers variant scoring
on two GPUs. Beyond Evo~2,
the case shows that operator agreement, model function and application agreement can disagree, and suggests a
cheap screen, untested elsewhere, for other FP8 checkpoints: rerun their tests accumulating the same operands
in fp32 (a pass does not rule dependence out). The study
concerns one model family, and whether training produced the dependence is not tested.

\begin{figure}[tbp]
\centering
\includegraphics[width=\textwidth,height=0.62\textheight,keepaspectratio]{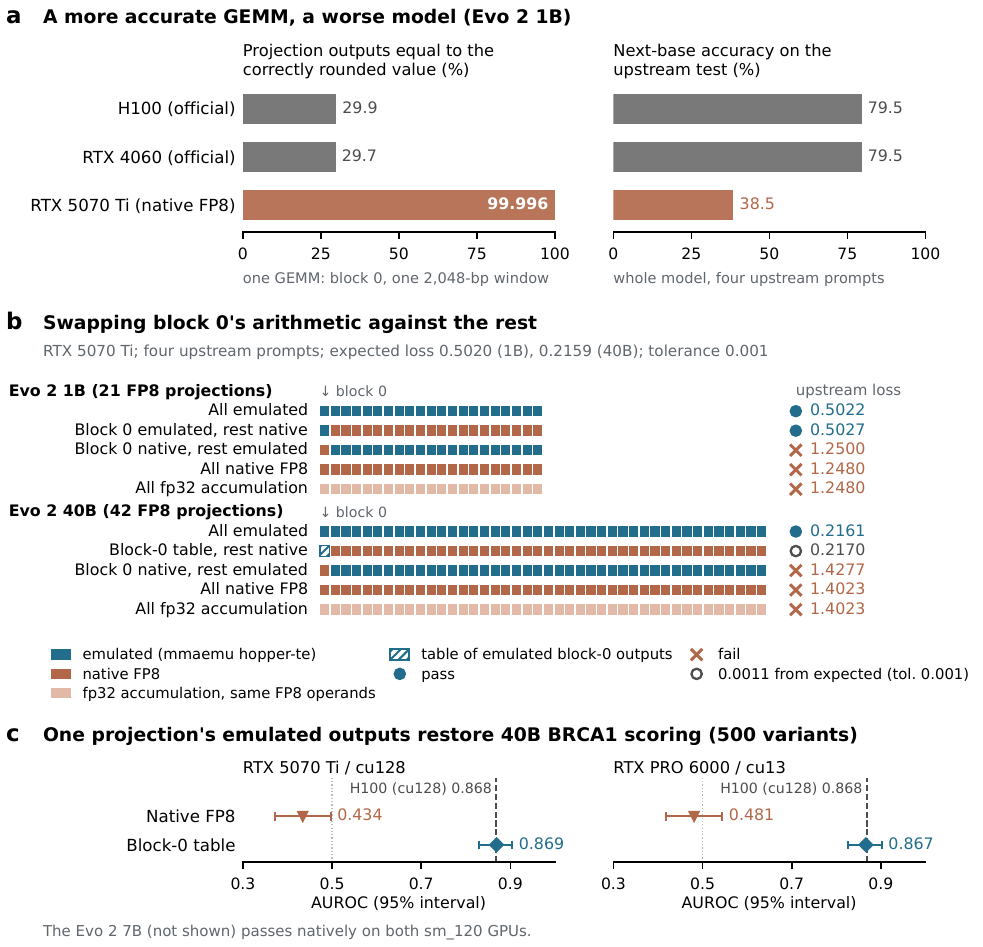}
\caption{\textbf{Arithmetic accuracy and checkpoint compatibility are different targets.}
\textbf{a},~Evo~2 1B. Left: share of the 11,796,480 outputs of block~0's FP8 projection on one 2,048-bp window
that equal the correctly rounded bf16 value (one GEMM, whose rows embed only four distinct tokens: 23,040 distinct dot products; other projections and windows were not analysed). Right:
next-base accuracy of the whole model over the four upstream test prompts (printed by the test; not its pass
criterion). The comparison is across devices: the reference GPUs ran the official Transformer Engine path,
whereas the RTX~5070~Ti's native FP8 projections called \code{torch.\_scaled\_mm}; panel~b's fp32 rows are the
within-device control.
\textbf{b},~Each square is one FP8 input projection, in model order; the first is block~0's. All runs used the
RTX~5070~Ti, with every other operation native. Emulated squares are live mmaemu \code{hopper-te}; the hatched
square is a lookup table of emulated \code{hopper-te} block-0 outputs; fp32 squares accumulate the same E4M3
operands in IEEE fp32. Filled circles pass the upstream forward test (within 0.001 of the expected loss),
crosses fail, and the open circle lies 0.0011 from the expected value, just outside the tolerance; with the other
40B projections in bf16 the table gives 0.2168, within it (Table~\ref{tab:runs}). Each loss is a mean of four bf16
per-prompt values, so distinct runs can print identical means (e.g.\ 1.2480).
\textbf{c},~40B AUROC on 500 \textit{BRCA1} variants (111 LoF), with conditional 95\% variant-bootstrap
intervals. Within each native/table pair the GPU, weights, software stack and cohort are fixed and only
block~0's FP8 projection changes; on the PRO~6000, native is the unmodified upstream path. The dashed line is the
H100 reference (cu128), so comparing the PRO~6000 (cu13) column with it changes both device and stack; the dotted
line marks 0.5. AUROCs are not ranked, and recovered variant scoring does not certify the forward or generation
tests.}
\label{fig:overview}
\end{figure}

\section{Results}

\subsection{A more accurate GEMM breaks two checkpoints}\label{sec:fail}

Two of the three tested checkpoints fail on both sm\_120 GPUs. We ran the published weights on the reference
GPUs and on an RTX~5070~Ti (sm\_120), varying only the FP8 input projections (21 GEMMs in the 1B, 42 in the
40B) and leaving all other computation native (Methods). The upstream forward test passes if the mean
next-base cross-entropy over four prompts lies within 0.001 of an expected value. With native FP8 (the GPU's own kernels), the
RTX~5070~Ti gave 1.2480 for the 1B and 1.4023 for the 40B, 2.5 and 6.5 times the expected 0.5020 and
0.2159; the H100 passed both.
The 40B predicted the next base correctly 25.7\% of the time, close to the 25\% of guessing among four bases,
against 91.6\% on the same GPU with mmaemu's emulation of the H100's TE path (\code{hopper-te}) and 91.7\% on
the H100. The 1B's accuracy roughly halved (Fig.~\ref{fig:overview}a), and it
failed the generation test (32.65\% identity against $68.0\pm3$). Running bf16 GEMMs, which change operand quantisation
as well as accumulation, failed (Table~\ref{tab:runs}). The 7B, whose input projections also use
FP8, passed natively (0.3477). An RTX~PRO~6000 running the unmodified upstream path under a separate CUDA~13 (cu13)
stack showed the same pattern (Table~\ref{tab:runs}; the 7B passed both tests), as did a public 1B report from
another RTX~5070~Ti \cite{evo2issue157}.

At the measured projection, the native GEMM was the more accurate one (Fig.~\ref{fig:overview}a). For
the 1B's first projection on one 2,048-bp window, the RTX~5070~Ti returned the correctly rounded bf16 value for
99.996\% of outputs, against about 30\% on either reference GPU, whose deviations lean towards zero. We then kept each projection's E4M3 operands, scales and output epilogue fixed and
accumulated the products in IEEE fp32 or, for the 1B, also in float64 rounded once to fp32. Both gave the native
loss of 1.2480 in the 1B, and fp32 accumulation gave the native 1.4023 in the 40B: more
accurate accumulation of the same operands, with no FP8 kernel involved, reproduces the failure
(Fig.~\ref{fig:overview}b). Emulating the H100's more accurate slow-accumulation mode (Methods) at block~0 also failed (1.2275), consistent with BioNeMo's note \cite{bionemo_evo2_docs}.

Exact agreement with one reference GPU was not needed either. The H100 and RTX~4060 differed on every one of
the 21 captured projections of this window, yet both supported the expected model behaviour (the RTX~4060 by
fp32 loss and AUROC); their tensor cores accumulate in blocks of different sizes
\cite{khattak2026accurate}. For the tested first-projection calls, a truncating chain retaining 25 fraction bits
matched the RTX~5070~Ti's native outputs bit for bit; this is not a general model of sm\_120.

The failure reached a scientific application. Following the upstream notebook, we scored a seeded subset of
500 \textit{BRCA1} variants, classified by a genome-editing assay \cite{findlay2018brca1} (111 loss-of-function,
LoF), by the change in mean sequence log-probability between
variant and reference windows (Table~\ref{tab:runs}). The 1B's area under the receiver operating characteristic
curve (AUROC) was 0.733 on the H100 and 0.738 on the RTX~4060, but 0.559 with native FP8. The 40B gave 0.868 on
the H100 and 0.434 natively, below 0.5 (95\% interval 0.373--0.497; we do not interpret the direction), with
per-variant scores essentially uncorrelated with the H100's (Pearson $-0.032$; 50.4\% same sign).

\begin{table}[tbp]
\centering
\caption{\textbf{Checkpoint behaviour under different FP8 projection implementations.}
The upper panel uses the reference GPUs and RTX~5070~Ti with the original cu128 (CUDA 12.8) stack; the lower panel uses
the PRO~6000 with cu13 (CUDA 13). Interventions are confined to originally FP8 projections; bf16 GEMMs also change operand quantisation.
Loss is the mean bf16 upstream-test loss; because each prompt's loss is a bf16 value, distinct runs often print identical means. AUROC has a 95\% variant-bootstrap interval. $r$ and sign are
Pearson correlation and same-sign percentage of per-variant $\Delta$ against the H100. Both panels use
the same 500 variants (111 LoF).
PRO--H100 comparisons change both device and software stack. Reference-pair and emulated AUROC intervals
overlap; no ranking is read.}
\label{tab:runs}
\footnotesize
\setlength{\tabcolsep}{2.6pt}
\begin{tabular}{@{}l r c r r r c r r@{}}
\toprule
 & \multicolumn{4}{c}{Evo~2 1B (500 variants, 111 LoF)} & \multicolumn{4}{c}{Evo~2 40B (500 variants, 111 LoF)\textsuperscript{g}} \\
\cmidrule(lr){2-5}\cmidrule(l){6-9}
Run & Loss & AUROC (95\% CI) & $r$ & Sign (\%) & Loss & AUROC (95\% CI) & $r$ & Sign (\%) \\
\midrule
H100, official & 0.5022 & 0.733 (0.673--0.792) & \multicolumn{2}{c}{ref.} & 0.2159 & 0.868 (0.829--0.904) & \multicolumn{2}{c}{ref.} \\ %
RTX~4060, official & n.r.\textsuperscript{a} & 0.738 (0.680--0.797) & 0.987 & 84.2 & \multicolumn{4}{c}{out of memory (8~GB)} \\ %
mmaemu \code{hopper-te} & 0.5022\textsuperscript{b} & 0.735 (0.675--0.793) & 0.988 & 86.0 & 0.2161 & 0.871 (0.833--0.905) & 0.998 & 97.4 \\ %
mmaemu \code{ada-te} & 0.5027 & 0.732 (0.672--0.791) & 0.986 & 84.2 & 0.2161 & 0.871 (0.833--0.905) & 0.998 & 96.2 \\ %
native FP8 & 1.2480 & 0.559 (0.501--0.615)\textsuperscript{c} & 0.058\textsuperscript{c} & 58.8\textsuperscript{c} & 1.4023 & 0.434 (0.373--0.497) & $-$0.032 & 50.4 \\ %
bf16 GEMMs & 1.3652 & 0.484 (0.419--0.545)\textsuperscript{c} & 0.065\textsuperscript{c} & 57.8\textsuperscript{c} & 1.4043 & 0.429 (0.367--0.497) & $-$0.010 & 50.2 \\ %
fp32 accumulation\textsuperscript{d} & 1.2480 & 0.556 (0.494--0.616) & 0.096 & 59.2 & 1.4023 & \multicolumn{3}{c}{not run} \\ %
float64 accumulation\textsuperscript{d} & 1.2480 & 0.527 (0.467--0.588) & 0.052 & 62.0 & \multicolumn{4}{c}{not run} \\ %
\midrule
\begin{tabular}[t]{@{}l@{}}block-0 table,\\ rest native FP8\textsuperscript{e}\end{tabular} & 0.5027 & 0.735 (0.676--0.794)\textsuperscript{c} & 0.986\textsuperscript{c} & 86.0\textsuperscript{c} & 0.2170 & 0.869 (0.830--0.904) & 0.997 & 97.4 \\ %
\begin{tabular}[t]{@{}l@{}}block~0 native,\\ rest \code{hopper-te}\textsuperscript{e}\end{tabular} & 1.2500 & \multicolumn{3}{c}{not run} & 1.4277 & \multicolumn{3}{c}{not run} \\ %
\midrule
\multicolumn{9}{l}{\textbf{RTX~PRO~6000 follow-up (cu13)}} \\
 & \multicolumn{4}{c}{1B (500 variants, 111 LoF)} & \multicolumn{4}{c}{40B (500 variants, 111 LoF)} \\
\cmidrule(lr){2-5}\cmidrule(l){6-9}
native FP8\textsuperscript{f} & 1.248 & 0.556 (0.494--0.618) & 0.070 & 60.4 & 1.402 & 0.481 (0.418--0.544) & $-$0.082 & 53.2 \\ %
block-0 table & n.r.\textsuperscript{f} & 0.721 (0.658--0.782) & 0.987 & 85.0 & n.r.\textsuperscript{f} & 0.867 (0.826--0.902) & 0.997 & 95.2 \\ %
\bottomrule
\end{tabular}

\smallskip
\begin{minipage}{\linewidth}
\scriptsize\raggedright
\textsuperscript{a}Not recorded; its fp32 loss is 0.5024. %
\textsuperscript{b}Equal to the H100's only through bf16 rounding of each prompt's loss: fp32 losses 0.5024 (\code{hopper-te}) and 0.5026 (H100); no window's logits are bit-identical. %
\textsuperscript{c}Computed for this paper, after the runs, from their archived scores. %
\textsuperscript{d}The same E4M3 operands and scales, dequantised exactly and accumulated in IEEE fp32 (TF32 off) or in float64 rounded once to fp32, then mmaemu's epilogue.
\textsuperscript{e}Block-0 table: block~0's FP8 projection reads a table of \code{hopper-te} outputs for all 512 token ids (plus zero pad rows), computed by mmaemu on the RTX~5070~Ti; all other FP8 projections run native FP8. The 40B's 0.2170 lies 0.0011 from the expected value, outside the tolerance; with the other projections in bf16 it gave 0.2168, within it (two bf16 rounding steps apart, on one of the four prompts). Block~0 native: only block~0's projection runs native FP8.
\textsuperscript{f}The PRO table's four-prompt forward loss was not included in the original run. Its AUROCs therefore
do not certify a passing forward test. The PRO native losses are printed to three decimals in the upstream log.
On the PRO~6000, native FP8 is the unmodified upstream Transformer Engine path; the patched path matched its
logits bit for bit on the gate window.
\textsuperscript{g}The 40B RTX~5070~Ti scores are from the later 500-variant follow-up; forward losses are from the earlier four-prompt tests.
\par
\end{minipage}
\end{table}

\subsection{Swapping one projection's arithmetic localises the dependence}\label{sec:block0}

mmaemu computes the scaled FP8
GEMM of a named reference platform, a specific H100 PCIe or RTX~4060 configuration, on another CUDA GPU
\cite{ding2026mmaemu}. Its profiles model the tensor-core accumulation chain, the library's split-K behaviour
and the output epilogue. It reproduces bit for bit all 168 captured H100 outputs of the 1B's 21 projections (two
output types, two accumulation modes, two APIs; Methods), and a
rule for combining split-K partial sums, fitted on an initial probe, matched all 2,416 split calls of a later
sweep at the splits the H100 used. Longer windows and all 40B projections rely on the validated model rather
than direct H100 captures. With either profile (\code{hopper-te}, or \code{ada-te} for the RTX~4060's TE path), both checkpoints
passed the upstream forward test on the RTX~5070~Ti (Table~\ref{tab:runs}).
Whole-model logits still differed from the reference GPUs', because convolutions, attention, bf16 GEMMs and
normalisation ran natively.

We then changed individual projections (Fig.~\ref{fig:overview}b) under a locally hashed, self-attested
analysis plan (PLAN\_C; Methods). With only block~0's
projection computed by \code{hopper-te} and the other 20 native, the 1B passed its forward test (0.5027); with
block~0 native and the other 20 emulated, it failed (1.2500). The 1B's block~23, whose H100 outputs shrink more
(least-squares gain against exact outputs 0.982, against 0.986 at block~0), stayed native in the passing configuration. Block~0's
projection sees the normalised embedding of a single token, so its outputs for all 512 token ids (plus zero pad
rows) fit in a lookup table, which matched live emulation on every call. With this table and every other
projection native, the 1B also passed the generation test (70.95\% against $68.0\pm3$).

The same projection mattered in the 40B. With block~0 native and the other 41 projections emulated, it failed
(1.4277). With block~0's table and every other projection native, it gave 0.2170, a deviation of 0.0011 from the
expected value against a tolerance of 0.001; with the others in bf16 it gave 0.2168, within tolerance
(Table~\ref{tab:runs}). In both checkpoints, native arithmetic at this
projection is enough to cause failure, and in the 40B restoring it alone comes close to restoring the forward
test. The fine-tuning project's block-23 localisation \cite{l1joseph_evo2ft} uses other criteria and GPUs.

\subsection{In the 1B, a slight shrinkage at the first projection passes and exactness fails}\label{sec:gemm}

\begin{figure}[tbp]
\centering
\includegraphics[width=0.95\textwidth]{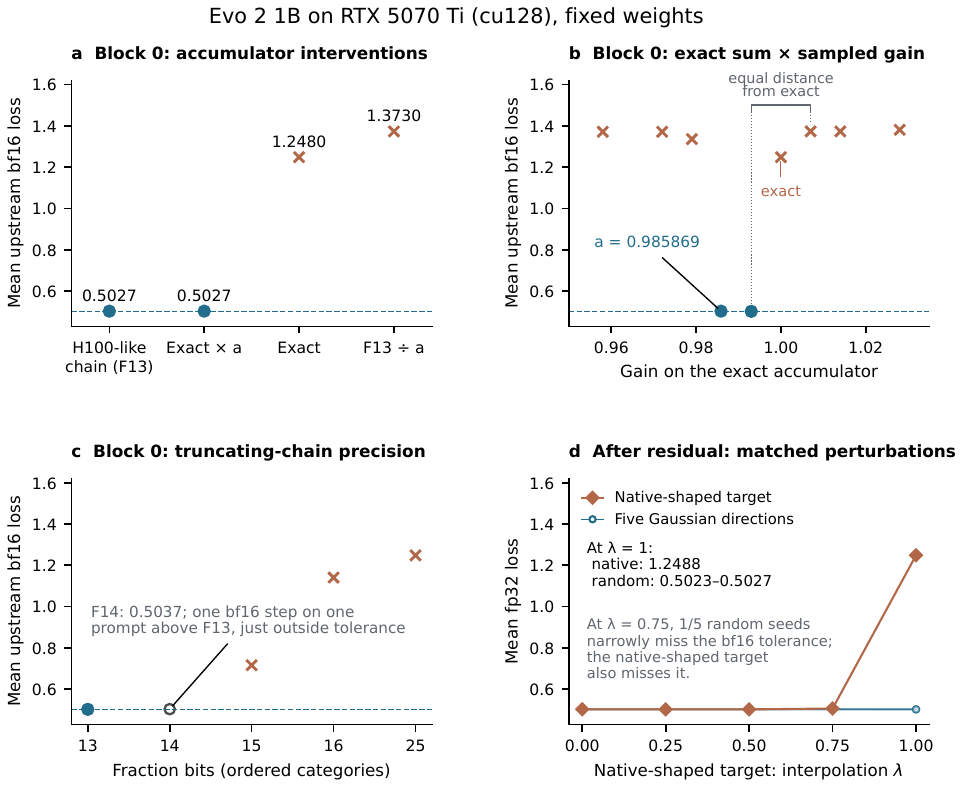}
\caption{\textbf{Local arithmetic and matched residual interventions in Evo~2 1B.}
All displayed losses use the RTX~5070~Ti and the four upstream forward-test prompts, in nats/base.
\textbf{a--c},~Only block~0's FP8 accumulator changes; operands, scales and the epilogue are common.
Loss is the upstream mean bf16 metric; the dashed line is the expected value, 0.501953125. Filled circles pass
its tolerance, crosses fail and the open circle marks a narrow miss outside the tolerance (same marker as Fig.~\ref{fig:overview}b).
\textbf{a},~The H100-like truncating chain (F13, equal to \code{hopper-te} bit for bit) and the exact
accumulator times the H100's measured gain ($a=0.985869$) pass; exact accumulation and the F13 accumulator divided
by $a$ fail. The gain is fitted to known arithmetic outputs, not loss, and is in-sample. These treatments
do not independently vary error pattern and scale; output rounding remains.
\textbf{b},~Nine sampled gains on the exact accumulator; no passing interval or optimum is inferred between points.
Gains 0.993 and 1.007 lie at equal distance from the exact output; the shrinkage passes and the enlargement fails.
\textbf{c},~Truncating-chain fraction bits are ordered categories with equal visual spacing.
F14's bf16 mean differs from F13's passing mean only by one bf16 rounding step on one prompt, which takes it
just outside the tolerance (its fp32 loss, 0.5027, is no higher than F13's, 0.5028); from 15 bits the loss rises sharply.
\textbf{d},~A separate intervention after block~0's residual addition and before the MLP, using the block-0 F13
path with the other FP8 projections native. The native-shaped target is
$\mathrm{bf16}(r_F+\lambda(r_N-r_F))$, where $r_F$ and $r_N$ are the block-0 F13 and native residuals.
At each amplitude, five Gaussian directions match the realised target
perturbation's per-token L2 norms within 1\% after bf16 rounding. One zero-amplitude reference is shared.
Each point is the mean \emph{fp32} loss across
four prompts; thin lines join discrete evaluations. At amplitude 0.75 one random direction narrowly misses the
\emph{bf16} test tolerance and the native-shaped target also misses it; only the native-shaped target at full amplitude gives
native-like loss (1.2488), while all five random directions retain low loss. These controls concern this node,
checkpoint and prompt set.}
\label{fig:gemm}
\end{figure}

At the 1B's block~0 we compared four accumulator treatments, all followed by the common epilogue
(Fig.~\ref{fig:gemm}a). The H100-like 13-bit truncating chain (F13, equal to \code{hopper-te} bit for bit) passed
at 0.5027, and exact accumulation failed at 1.2480. Multiplying the exact accumulator by the H100's
least-squares gain at this projection, $a=0.985869$ (a 1.4\% shrinkage towards zero, close to F13's own gain),
passed at 0.5027, although its outputs differ from the reference chain's; dividing the reference accumulator by
$a$, which keeps the truncation pattern but removes the shrinkage, failed at 1.3730. The gain was fitted to
arithmetic outputs on known inputs, not to loss, and is in-sample: a diagnostic, not a correction. In the 40B,
the analogous controls, with a gain fitted in-sample to emulated outputs on the test's own tokens (0.938), gave 0.2173 for the scaled exact sum (0.0013 from expected, outside tolerance) and 1.4336 for
the divided reference chain.

Only two of nine sampled gains on the exact accumulator passed, about 0.986 and 0.993; larger shrinkages of
2.1--4.2\% (0.979, 0.972, 0.958), 1.000 and three enlargements failed
(Fig.~\ref{fig:gemm}b). Gains 0.993 and 1.007 lie at the same distance from the exact output, yet the shrinkage
passed (0.5020) and the enlargement failed (1.3730). Nor did GEMM-output distance from reference outputs decide.
Random multiplicative noise on the H100-like accumulator, which kept its gain, passed for all three seeds at two
such distances: native's (0.016) and the failing 0.979 gain's (0.010). Gains 1.000 and 0.972 at the first distance
and 0.979 at the second failed. Downstream, this noise shrinks and native's difference grows
(post hoc). Raising the truncating chain's precision gave losses of
0.5027, 0.5037 (narrowly outside tolerance), 0.7163 and 1.1406 at 13--16 fraction
bits and 1.2480 at 25 bits (Fig.~\ref{fig:gemm}c), and 13 bits with round-to-nearest instead of truncation failed
(1.1924). Less accuracy alone therefore does not restore the model, and shrinking and enlarging differ: in these sparse
samples the outcome followed a small shrinkage near the H100's rather than GEMM-output distance from the exact or
reference outputs.

\subsection{Matched-size random changes after block~0 did not collapse the model}\label{sec:downstream}

In captures from a separate 35-run study under a locally hashed plan, block~0's residual addition largely
cancels its branch output: across it, the native-minus-F13 discrepancy kept its absolute size but grew
12.5--13.4-fold in relative size, so the sum is about that much smaller in norm than the branch. The study
asked whether size alone explains the collapse, perturbing the residual stream after this addition and before block~0's MLP
(Fig.~\ref{fig:gemm}d). Injecting the full native-minus-F13 residual difference gave a mean fp32 loss of 1.2488.
Five Gaussian directions whose realised per-token L2 norms matched it within 1\% gave 0.5023--0.5027 and passed
the test's bf16 criterion; at amplitude 0.75 ($\lambda$, the injected fraction), one
random direction narrowly missed it and the native-shaped target also missed it. Splitting the observed branch
difference into components parallel and orthogonal to the reference branch gave fp32 losses of 0.5094 and
0.5033 separately, both missing the bf16 criterion but far below their combination's 1.2488. At this node and on these prompts,
the structured native difference collapsed the model while matched-norm random changes did not; direction,
channel structure and cross-token correlation were not separated.

\subsection{One projection's emulated outputs restore variant scoring on two GPUs}

For the 1B's 500 variants, full \code{hopper-te} emulation and the H100 had score-change Pearson correlation
0.988 and 86.0\% sign agreement, compared with 0.987 and 84.2\% between the two reference GPUs; paired bootstrap
differences included zero and AUROC intervals overlapped, which shows agreement, not equivalence or rank. The
block-0 table with all other projections native gave AUROC 0.735, Pearson 0.986 and 86.0\% sign
agreement against the H100, not distinguishable from the reference pair at this sample size.

\begin{figure}[tbp]
\centering
\includegraphics[width=\textwidth,height=0.76\textheight,keepaspectratio]{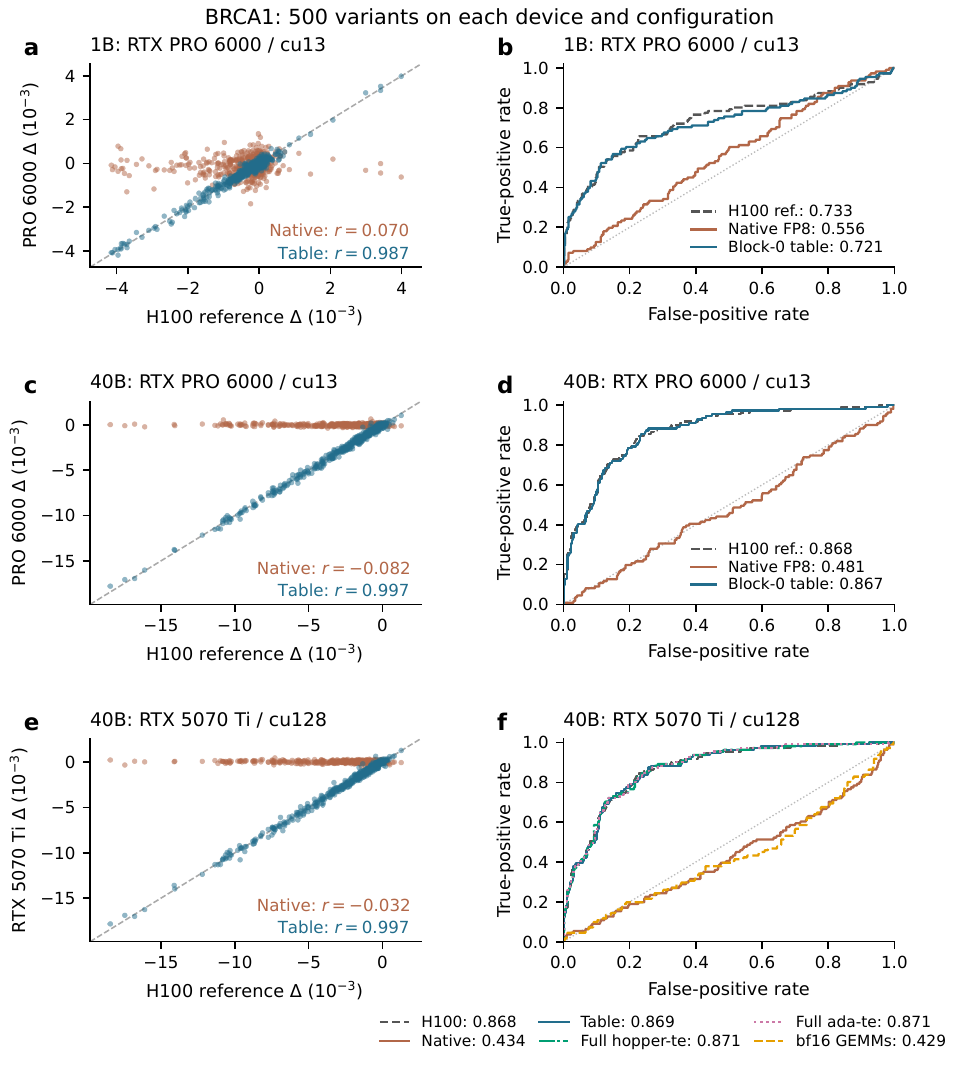}
\caption{\textbf{\textit{BRCA1} variant-effect scoring on two sm\_120 GPUs.}
All panels use the same 500 \textit{BRCA1} variants.
\textbf{a,b},~1B on the PRO~6000 (cu13). \textbf{c,d},~40B on the PRO~6000 (cu13).
\textbf{e,f},~40B on the RTX~5070~Ti (cu128).
Left panels compare native FP8 and the block-0 table of emulated \code{hopper-te} outputs with the archived H100
reference; each point is one variant, dashed lines mark equality and $r$ is Pearson correlation.
Only block~0's FP8 projection changes between native and table runs; the remaining model runs natively.
Right panels show ROC curves; legend values are AUROCs. Panel~f, whose legend sits below it, also includes full
\code{hopper-te} and \code{ada-te} emulation and bf16 GEMMs. Table~\ref{tab:runs} gives 95\% variant-bootstrap
intervals; no ranking is inferred among the reference and emulated AUROCs, and AUROC agreement alone is not
a model-level equivalence test. The H100 reference used cu128, as did the RTX~5070~Ti; PRO--H100 comparisons
change both device and software. $\Delta$ is variant minus reference mean sequence log-probability; ROC curves use
$-\Delta$ with LoF positive and functional/intermediate variants negative.}
\label{fig:variants}
\end{figure}

For the 40B on the RTX~5070~Ti, the table raised AUROC from 0.434 to 0.869 (paired difference 0.435, 95\%
interval 0.360--0.507), with Pearson correlation 0.997 and 97.4\% sign agreement against the H100
(Figs.~\ref{fig:overview}c and \ref{fig:variants}e,f); full \code{hopper-te} emulation gave 0.998 and 97.4\%
(no 40B reference-pair band exists: the RTX~4060 ran out of memory). The table's forward loss of 0.2170
still misses tolerance (Section~\ref{sec:block0}): application agreement is distinct from forward-test
compatibility.

Within the PRO~6000 configuration (cu13), the same intervention raised AUROC from 0.556 to 0.721 for the 1B and
from 0.481 to 0.867 for the 40B on the same 500 variants (Fig.~\ref{fig:variants}a--d); the 40B's paired
improvement was 0.385 (0.306--0.463). The 40B's per-variant correlation with the earlier H100 reference rose from
$-0.082$ to 0.997 (Table~\ref{tab:runs}), but that comparison changes both device and software stack, whereas the
within-PRO intervention holds both fixed.

\section{Discussion}

For fixed weights, the more accurate GEMM was the incompatible one, as BioNeMo observed on Hopper
\cite{bionemo_evo2_docs}. These experiments trace this change to one projection, to a candidate property in the
1B, and back to recovery with the weights unchanged. The lesson
is not that less accurate arithmetic is safer: a larger shrinkage and round-to-nearest at reference precision
failed as surely as exact sums. Rather, a fixed checkpoint can depend on specific properties of its reference
arithmetic. We do not estimate how common such dependence is among low-precision models; even within Evo~2, the
7B passes.

Validation should separate operator agreement, model function and application agreement, which disagreed here.
The two reference GPUs differ in every projection's bits yet both support the expected behaviour; full emulation
passes the forward tests with logits that differ from the H100's; the 40B table restores AUROC while missing the
forward tolerance. Repeated runs of the tested calls were bit-identical on each platform, and on the tested
fixed-operand calls the two sm\_120 GPUs matched bit for bit under the same cu13 stack, whereas the RTX~5070~Ti's native outputs
for the first projection differed from both reference GPUs': determinism is not compatibility. The two
reference GPUs also agreed on the sign of only 84.2\% of 1B variant effects; all disagreements involved effects
below four times the mean cross-GPU difference in $\Delta$ (computed for this paper). Scientific analyses should retain execution provenance and read small score changes against measured
platform variation.

Evo~2's upstream tests exposed this dependence.
Releases could add application reference outputs and tolerances, together with the GPU configuration, software
stack and effective accumulation mode. Libraries could expose the mode actually used: cuBLAS documents its
fast-accumulation flag as taking effect only on Ada and Hopper \cite{nvidia_cublas_12_8}, and on the
RTX~5070~Ti the flag changed none of the tested outputs. Substituting emulated reference GEMMs one projection at a time could then
localise such dependence; like the fp32 screen (Introduction), it is untested beyond Evo~2. Where a sensitive layer's input depends only on the token id, its
reference outputs can be tabulated: on the RTX~5070~Ti, an 8,192-bp 1B forward pass took 16\% longer with the
table than natively (medians; table construction excluded), whereas full emulation took 2.2 times as long; with
native FP8 elsewhere, the 40B table missed the forward tolerance. Fine-tuning offers another route, with native kernels and changed weights
\cite{bionemo_evo2_docs}; BioNeMo's original 1B working on a B300, under another stack and evaluation, is an
important boundary case that these experiments do not explain.

Training could compensate for an arithmetic effect whose later removal changes the model's function: if
$r=x+\alpha wx$ (input $x$, GEMM gain $\alpha$, weight $w$) is fitted to a small target $\epsilon x$, replacing
$\alpha$ by 1 changes $r$ by $(1-\alpha)(\epsilon-1)x/\alpha$, amplifying the GEMM output's relative change
$(1-\epsilon)/\epsilon$-fold. Block~0's cancelling residual addition (Section~\ref{sec:downstream}) is such a regime, but random changes of the same size, and either branch component alone, did not collapse the model; the construction, which models neither
FP8 arithmetic nor Evo~2's training, is a candidate explanation only. BioNeMo reports that compatibility can move
with the weights: a further-trained 40B fine-tune degrades on ordinary Hopper FP8 but not with more accurate accumulation
or on a B300 \cite{bionemo_evo2_docs}. Establishing a training origin requires controlled training or
fine-tuning with crossed evaluation under both arithmetic implementations.

\paragraph{Limitations.}
We tested one model family on two sm\_120 GPUs; the 20B, B200 and B300 were not run, and the 7B has no reference-GPU comparison. Reference evidence is one H100 PCIe and one RTX~4060
configuration. The PRO follow-up changes software as well as hardware relative to the original experiments. There are no direct
H100 captures of 40B projections or tests above 8,192 projection rows. The 40B table misses the forward
tolerance on the RTX~5070~Ti, the PRO table's forward loss was not recorded, and 40B generation was not tested
under emulation or with the table. Biology is one assay and one 500-variant cohort; bootstrap intervals are
conditional on the recorded runs and model neither overlapping-window dependence nor run-to-run variation. The
gain is in-sample, and no rescaling correction was tested. Mechanism controls use four previously studied
prompts. mmaemu does not
support bias, row-wise scales, autograd or CUDA graphs.

\section{Methods}

\paragraph{Models and software.}
We used the Evo~2 checkpoints \code{evo2\_1b\_base} and \code{evo2\_40b} \cite{brixi2026evo2}, and \code{evo2\_7b} (PLAN\_C and PRO~6000, below). The 1B has 21 FP8 input projections with $K = 1{,}920$ and $N = 5{,}760$; the 40B has 42, with $K = 8{,}192$
and $N = 24{,}576$. Original reference runs used evo2 0.6.0, vtx 1.1.0
\cite{vortex}, torch 2.7.1+cu128 \cite{ansel2024pytorch2} and Transformer Engine 2.3.0; the RTX~5070~Ti ran the same
stack without Transformer Engine, its native FP8 projections calling \code{torch.\_scaled\_mm}. The original Evo~2 runs with
mmaemu \cite{ding2026mmaemu} predate
its 0.1.0 tag (\code{ada-te}: development build 0.1.0.dev0, 14--15 October; \code{hopper-te}: commit fa0fd90, 16 October); with
the released code, the upstream tests (1B with both profiles; 40B with \code{hopper-te}) and 40 \textit{BRCA1}
windows gave bit-identical outputs. Later runs used the
release.

\paragraph{Platforms.}
The reference platforms are an NVIDIA H100 PCIe (sm\_90, 114 SMs, driver 570.148.08) and a GeForce RTX~4060 (sm\_89,
24 SMs, driver 572.16), both with cuBLASLt 12.8.3.14. The H100 was rented four times on 14--16 October 2026. The RTX~5070~Ti (sm\_120, 70 SMs, 16~GB,
driver 610.88) runs mmaemu but is not a reference platform.

\paragraph{Execution modes.}
On the RTX~5070~Ti, bf16 mode runs the original weights in a bf16 GEMM, and mmaemu uses kernels h4
(\code{hopper-te}) and v4 (\code{ada-te}). The 40B, too large
for the GPU, runs one block at a time from host memory (offload). On the 1B, offload gave the resident model's logits
exactly on every test window checked. The official 40B runs,
and the H100's 1B test losses, ran the upstream blocks one at a time with Transformer Engine; on the 1B this gave the
upstream model's logits bit for bit on both reference GPUs.

\paragraph{PRO~6000 follow-up.}
The follow-up used an RTX~PRO~6000 Blackwell Server Edition (sm\_120, 188 SMs, 96~GB), driver 580.178.04,
torch 2.14.1+cu130, cuBLASLt 13.1.1 and Transformer Engine 2.20.2, with the same Evo~2/vtx versions and
released mmaemu tree. Its plan and package hashes preceded GPU measurements; the plan was informed by earlier
results and was not externally preregistered. All three checkpoints passed full-file SHA-256 checks before loading. Models stayed resident. For operator comparisons, the RTX~5070~Ti ran the same torch and cuBLASLt.

The same exported cohort supplied 435 reference and 500 variant windows to the upstream scoring API,
with batch size one and no BOS. The table substitution was gated by bit-identical patched-native and official
logits on one prespecified reference window, then matched live emulation on every substituted block-0 call.
New AUROC intervals and paired differences use the existing variant-bootstrap procedure, computed post hoc.

The unmodified forward and generation tests ran in fresh processes. After the original resident 40B generation
test exhausted memory during Hyena FFT cache prefill, a dated supplement reran it through the public API with
\code{force\_prompt\_threshold=128}, keeping weights, prompts, seed and scoring but changing prefill scheduling;
the native FP8 40B on the PRO~6000 reached 23.95\% identity against $91.15\pm3$ expected. The supplement is reported separately
from the original matrix, which completed 13 of 14 application cells.

\paragraph{Evaluation and statistics.}
The upstream test (\code{test\_evo2.py}) averages the bf16 next-base cross-entropies (natural logarithm) of four
prompts and passes within $10^{-3}$ of the expected loss (1B 0.501953125; 7B 0.3476563; 40B 0.2159424). It also
prints next-base accuracy (argmax agreement), which is not part of its pass rule; guessing among four bases gives
25\%. The generation test greedily continues the first half of each prompt for 500 bases and passes within 3
points of the expected identity with the true continuation (1B 68.0\%, 7B 89.25\%, 40B 91.15\%).
We followed the upstream notebook's \textit{BRCA1} scoring procedure on a subset selected by
\code{sample(n=500, random\_state=0)}: 500 of 3,893 variants, including 111 LoF variants.
These AUROCs are not comparable with full-set values. Windows of 8,192~bp are centred on the
variant, and $\Delta$ is the variant window's mean next-base log-probability minus the reference window's. The AUROC
of $-\Delta$ for LoF has a 95\% percentile bootstrap interval (2,000 unstratified variant resamples; NumPy
\code{default\_rng(0)}). Functional and intermediate variants form the negative class. Paired comparisons use resamples shared across runs.
These intervals are conditional on the recorded inference runs and do not quantify uncertainty across genomic
regions, assays or GPU runs.

A later cu128 follow-up extended five RTX~5070~Ti 40B configurations from 100 to 500 variants on
20--21 October 2026. The first-100 outcomes were known; inputs and scripts were locally hashed before measurement.
Each configuration rescored 935 windows. All 198 first-100 window scores matched
their historical files bit for bit in every configuration; all 935 table calls matched live \code{hopper-te}.

\paragraph{Arithmetic model.}
mmaemu models a tensor-core FP8 GEMM as chains of block fused multiply-adds, each adding a block of products to
an incoming accumulator in one operation \cite{blanchard2020bfma,khattak2026accurate,matlab_tensor_core,fasi2021numerical},
with blocks of 16 (\code{ada}) or 32 (\code{hopper}) products. Each block aligns its products and incoming
accumulator to a common exponent, truncates the aligned terms towards zero to 13 fraction bits, sums them exactly,
then truncates the result to 13 fraction bits relative to its own leading bit. Slow accumulation sums 128-element
chains in round-to-nearest fp32. Split-K deals 64-wide $k$-tiles to the splits in turn on the RTX~4060. On the H100, each
split takes contiguous 128-wide tiles and the partial sums are added serially. The epilogue computes
$\mathrm{bf16}(\mathrm{RN}_{32}(\mathit{acc}\cdot\mathrm{RN}_{32}(s_a s_b)))$. For the 168 H100 comparisons,
full outputs were stored for three of the 21 projections and the first 256 rows for the other 18.

\paragraph{Additional runs for this paper.}
Runs A7, A9, A11 and A12 on the RTX~5070~Ti were specified, with their criteria and readings, in a plan hashed (self-attested) on 17
October 2026 before any of them started; all use tag \code{v0.1.0}.
\emph{A7} recomputed the first 1B projection natively from the recorded operands; as a gate, it reproduced the
recorded difference count. It then compared all outputs with the exact product, accumulated in
float64 (exact at $K = 1{,}920$), scaled and rounded once to bf16.
\emph{A9} (accumulation control) dequantises the E4M3 operands exactly, accumulates them in IEEE fp32 (TensorFloat-32
off) or in float64 rounded once to fp32, and applies mmaemu's epilogue. An amendment, hashed before the A9 runs, adds the 40B upstream test
with fp32 accumulation; a smoke test of both modes on two 1B windows had finished about one minute earlier, and
whether its output had been seen is not recorded. A second amendment dropped a planned 40B \textit{BRCA1} run before
it started. \emph{A11} timed native \code{torch.\_scaled\_mm} and kernels h4 and v4 with CUDA events on synthetic
operands of Evo~2 GEMM shapes; full emulated calls took 21.8--27.7 (h4) and 23.0--28.5 (v4) times
as long as native.
\emph{A12} rescored 40 windows of the \code{hopper-te} 1B \textit{BRCA1} pass, which had run through offload, on the
resident model.

\paragraph{Localisation runs (PLAN\_C).}
The single-projection runs of Sections~\ref{sec:fail}--\ref{sec:block0}, the runs of Section~\ref{sec:gemm}, the block-0 tables,
the generation test and the 7B were specified, with classes, gates and readings, in an analysis plan (PLAN\_C) frozen by SHA-256 on 18 October 2026 at
16:20 PDT, before any of its GPU runs. The hash is self-attested, not deposited with a third party before the runs, and the plan
text did not change. A ten-entry deviation log records re-hashed code fixes and run-setting changes, errata, an
analysis note, amendments for another plan and run-order changes; the timing step, initially deferred, completed on
19 October with the frozen command. The analysis scripts were not frozen before the runs; the \textit{BRCA1} statistics script was
written after the scores existed. Classes: strict, the upstream pass criterion; broken, above 1.0. The block-0
diagnostics, not fixes, were: the exact sum times $g$, or (Ea) times the H100's least-squares gain against it,
$a = 0.985869$, fitted on one of six evaluated windows (in-sample); \code{hopper-te}'s accumulator divided by $a$
(Ha) or times $1+\sigma z$ ($z$ standard normal per output, fixed per distinct input row; $\sigma = 0.016$ or
0.0104; three seeds); and truncating Hopper-structured chains with 13--16 or 25 fraction bits (13 equals
\code{hopper-te}, 25 native, bit for bit) or 13 bits rounded to nearest. The 40B analogues Ea40 and Ha40 use a gain
fitted likewise to \code{hopper-te}'s block-0 outputs over the test's own token rows. Distances are
token-weighted relative L2 distances, normalised by the exact outputs; a post hoc CPU recomputation agreed. Cells were grouped after the runs by precedent recorded before them (the same or an
analogous observation). Retests: for the 1B, the test losses of native, bf16 and full emulation, the block-0 localisation and
tables, the exact sum ($g = 1$, equal to float64 accumulation), the 13- and 25-bit chains (equal to \code{hopper-te} and native; that the 25-bit chain equals native bit for bit, gate G3, was new), native's generation failure (issue~\#132), the fast-accumulation flag (another software environment),
and block~23, slow accumulation and the table's \textit{BRCA1} scores (exploratory runs below); for
the 40B, native, full emulation and the table; for the 7B, native (issue~\#157; an exploratory
bf16 run). All other cells had no recorded precedent, including the scaled and perturbed block-0 outputs
($g \neq 1$, Ea, Ha, noise), the 14--16-bit and round-to-nearest chains, the fp32 residual additions, the generation
tests of the table and of full emulation, and the other 40B and 7B cells. The 7B also passed with full emulation
(0.3481) and the table (0.3477); the 1B's fp32 residual additions changed no class.

\paragraph{Matched residual interventions.}
A subsequent locally hashed plan, informed by PLAN\_C and not externally preregistered, specified 35
fresh-process 1B cells using the same cu128 stack and four prompts. We captured the actual biased branch $h$ and bf16
residual sum $r$ on the local F13 and native paths, with the other 20 FP8 projections native.
Residual replay and decomposition endpoints reproduced baseline logits bitwise. Five fixed Gaussian seeds were
evaluated at amplitudes 0.25, 0.5, 0.75 and 1, with a shared zero-amplitude reference. Realised post-bf16 per-token
perturbation norms matched the corresponding native-shaped injection within 1\% for every token (maximum
relative mismatch 0.166\%). Neither labels nor losses entered matching. Branch decomposition used one inner product
per complete prompt. The relative capture ratio of native-minus-F13 discrepancies is
$(\lVert\delta r\rVert/\lVert r\rVert)/(\lVert\delta h\rVert/\lVert h\rVert)$.

\paragraph{Exploratory runs.}
Runs on the RTX~4060 on 10--11 October 2026, before mmaemu's first commit and without criteria fixed in advance,
had shown most retested effects. A quantise--dequantise bf16 stand-in for block~0's FP8 GEMM gave a high 1B test
loss (1.081), whereas stand-ins for block~23 or the 19 others left it low. A table of block~0's Transformer Engine
outputs with bf16 elsewhere passed the 1B test, gave a 40B fp32 loss of 0.2165 and gave 1B \textit{BRCA1} $\Delta$
correlating at 0.988 with the official RTX~4060 run. With all 21 projections accumulated in
fp32, or in \code{torch.\_scaled\_mm} with fast accumulation off, the 1B gave high fp32 losses, but not with the
table and fp32 accumulation elsewhere; fast accumulation on matched Transformer Engine in every recorded digit. The
fp32 runs' code is not preserved, and the \code{torch.\_scaled\_mm} runs' script version is not established.

\section*{Code and data availability}
mmaemu 0.1.0 is free software under the Apache License 2.0, at \url{https://github.com/astding/mmaemu}, on PyPI as
\code{mmaemu}. The v0.1.0 release is archived at \url{https://zenodo.org/records/23440103}
(version DOI \href{https://doi.org/10.5281/zenodo.23440103}{10.5281/zenodo.23440103}) \cite{ding2026mmaemu}. Documentation:
\url{https://astding.github.io/mmaemu/}. Repository paths refer to commit \code{ab37ab1}, which differs from the tag
\code{v0.1.0} only in documentation and citation metadata.
The release's validation logs under \code{docs/validation\_logs/} contain the original hardware-validation and Evo~2
inference logs, result files and per-variant \textit{BRCA1} scores. \code{docs/validation.md} summarises them. The official RTX~4060 Evo~2 1B
outputs are in \code{examples/data/}, and \code{examples/evo2\_blackwell.ipynb} and \code{examples/evo2\_offload.py}
run Evo~2 with mmaemu. Continuous integration replays committed subsets of the RTX~4060 and H100 recordings.
Larger hardware recordings and later datasets, including runs A7, A9, A11 and A12, PLAN\_C with its deviation log,
the matched residual interventions, the RTX~PRO~6000 and RTX~5070~Ti 40B 500-variant follow-ups, the exploratory
RTX~4060 probes, a ledger tracing each reported number to its source file, post hoc analyses and figure scripts, are
deposited with SHA-256 manifests at Zenodo (DOI \href{https://doi.org/10.5281/zenodo.23740390}{10.5281/zenodo.23740390}).
Evo~2 checkpoints are the Arc Institute's releases on Hugging Face; the \textit{BRCA1} data are those of Findlay et
al.\ \cite{findlay2018brca1}, as read by the evo2 notebook.

\section*{Competing interests}
The authors declare no competing interests.

\section*{Funding}
This work received no funding. Both authors worked on personal time, with personal equipment and no institutional
resources; GPU rental time was paid for personally by X.D.

\section*{Author contributions}
X.D.\ and Z.Q.\ conceived and designed the study, wrote the software, ran the experiments and analyses, and wrote
the manuscript. Both authors reviewed and approved the final manuscript.
AI assistance is described below.

\section*{Acknowledgements}
The authors thank Faizan A.\ Khattak and Mantas Mikaitis for the MATLAB Tensor Core toolbox
\cite{matlab_tensor_core,khattak2026accurate}, from whose models mmaemu's block model and presets are translated and
whose validation vectors are part of mmaemu's tests. The authors thank the Arc Institute and the Evo~2
collaborators for their substantial contribution to genomic modelling and for publicly sharing model weights,
inference software, tests and evaluation notebooks. These resources made the present investigation possible.
The authors also thank the community contributors who documented their observations in evo2 issue~\#157.
This work and mmaemu are independent: they are not affiliated with, funded by or endorsed by NVIDIA Corporation,
the Arc Institute, X.D.'s employer or Emory University. Z.Q.\ contributed independently, in a personal capacity. NVIDIA, GeForce, RTX, CUDA, cuBLAS, Hopper, Blackwell, H100 and Transformer
Engine are trademarks of NVIDIA Corporation.

\section*{Use of AI assistance}
Claude (Anthropic) and Codex (OpenAI) assisted with developing and debugging computational code, planning and executing
experiments, analysing recorded outputs, preparing figures, and drafting and editing the manuscript.
Both authors take full responsibility for the manuscript's accuracy, integrity
and originality.

\bibliographystyle{unsrtnat}
\bibliography{refs}

\begin{thebibliography}{24}
\providecommand{\natexlab}[1]{#1}
\providecommand{\url}[1]{\texttt{#1}}
\expandafter\ifx\csname urlstyle\endcsname\relax
  \providecommand{\doi}[1]{doi: #1}\else
  \providecommand{\doi}{doi: \begingroup \urlstyle{rm}\Url}\fi

\bibitem[Micikevicius et~al.(2023)Micikevicius, Oberman, Dubey, Cornea,
  Rodriguez, Bratt, Grisenthwaite, Jouppi, Chou, Huffman, Schulte, Wittig,
  Jani, and Deng]{ofp8_2023}
Paulius Micikevicius, Stuart Oberman, Pradeep Dubey, Marius Cornea, Andres
  Rodriguez, Ian Bratt, Richard Grisenthwaite, Norm Jouppi, Chiachen Chou,
  Amber Huffman, Michael Schulte, Ralph Wittig, Dharmesh Jani, and Summer Deng.
\newblock {OCP} 8-bit floating point specification ({OFP8}), revision 1.0.
\newblock Specification, Open Compute Project, 2023.
\newblock URL
  \url{https://www.opencompute.org/documents/ocp-8-bit-floating-point-specification-ofp8-revision-1-0-2023-12-01-pdf-1}.
\newblock Approved 20 June 2023; document version of 1 December 2023.

\bibitem[Khattak and Mikaitis(2026)]{khattak2026accurate}
Faizan Khattak and Mantas Mikaitis.
\newblock Accurate models of {NVIDIA} tensor cores.
\newblock \emph{ACM Transactions on Architecture and Code Optimization},
  23\penalty0 (3):\penalty0 1--25, 2026.
\newblock \doi{10.1145/3830409}.

\bibitem[{NVIDIA}(2025{\natexlab{a}})]{nvidia_cublas_12_8}
{NVIDIA}.
\newblock {cuBLAS} library documentation ({CUDA Toolkit} 12.8),
  2025{\natexlab{a}}.
\newblock URL
  \url{https://docs.nvidia.com/cuda/archive/12.8.0/cublas/index.html}.
\newblock Last updated 23 January 2025.

\bibitem[Monniaux(2008)]{monniaux2008}
David Monniaux.
\newblock The pitfalls of verifying floating-point computations.
\newblock \emph{ACM Transactions on Programming Languages and Systems},
  30\penalty0 (3):\penalty0 1--41, 2008.
\newblock \doi{10.1145/1353445.1353446}.

\bibitem[Yuan et~al.(2025)Yuan, Li, Ding, Xie, Li, Zhao, Wan, Shi, Hu, and
  Liu]{yuan2025nondeterminism}
Jiayi Yuan, Hao Li, Xinheng Ding, Wenya Xie, Yu-Jhe Li, Wentian Zhao, Kun Wan,
  Jing Shi, Xia Hu, and Zirui Liu.
\newblock Understanding and mitigating numerical sources of nondeterminism in
  {LLM} inference.
\newblock In \emph{Advances in Neural Information Processing Systems},
  volume~38, pages 169819--169851, 2025.
\newblock \doi{10.52202/085713-5653}.

\bibitem[{DeepSeek-AI}(2024)]{deepseekai2024v3}
{DeepSeek-AI}.
\newblock {DeepSeek-V3} technical report.
\newblock arXiv:2412.19437, 2024.
\newblock arXiv v2 (18 Feb 2025).

\bibitem[Brixi et~al.(2026)Brixi, Durrant, Ku, Naghipourfar, Poli, Sun,
  Brockman, Chang, Fanton, Gonzalez, King, Li, Merchant, Nguyen, Ricci-Tam,
  Romero, Schmok, Taghibakhshi, Vorontsov, Yang, Deng, Gorton, Nguyen, Wang,
  Pearce, Simon, Adams, Amador, Ashley, Baccus, Dai, Dillmann, Ermon, Guo,
  Herschl, Ilango, Janik, Lu, Mehta, Mofrad, Ng, Pannu, R{\'e}, {St. John},
  Sullivan, Tey, Viggiano, Zhu, Zynda, Balsam, Collison, Costa,
  Hernandez-Boussard, Ho, Liu, McGrath, Powell, Pinglay, Burke, Goodarzi, Hsu,
  and Hie]{brixi2026evo2}
Garyk Brixi, Matthew~G. Durrant, Jerome Ku, Mohsen Naghipourfar, Michael Poli,
  Gwanggyu Sun, Greg Brockman, Daniel Chang, Alison Fanton, Gabriel~A.
  Gonzalez, Samuel~H. King, David~B. Li, Aditi~T. Merchant, Eric Nguyen, Chiara
  Ricci-Tam, David~W. Romero, Jonathan~C. Schmok, Ali Taghibakhshi, Anton
  Vorontsov, Brandon Yang, Myra Deng, Liv Gorton, Nam Nguyen, Nicholas~K. Wang,
  Michael~T. Pearce, Elana Simon, Etowah Adams, Zachary~J. Amador, Euan~A.
  Ashley, Stephen~A. Baccus, Haoyu Dai, Steven Dillmann, Stefano Ermon, Daniel
  Guo, Michael~H. Herschl, Rajesh Ilango, Ken Janik, Amy~X. Lu, Reshma Mehta,
  Mohammad R.~K. Mofrad, Madelena~Y. Ng, Jaspreet Pannu, Christopher R{\'e},
  John {St. John}, Jeremy Sullivan, Joseph Tey, Ben Viggiano, Kevin Zhu, Greg
  Zynda, Daniel Balsam, Patrick Collison, Anthony~B. Costa, Tina
  Hernandez-Boussard, Eric Ho, Ming-Yu Liu, Thomas McGrath, Kimberly Powell,
  Sudarshan Pinglay, Dave~P. Burke, Hani Goodarzi, Patrick~D. Hsu, and Brian~L.
  Hie.
\newblock Genome modelling and design across all domains of life with {Evo} 2.
\newblock \emph{Nature}, 652\penalty0 (8112):\penalty0 1349--1361, 2026.
\newblock \doi{10.1038/s41586-026-10176-5}.

\bibitem[{Arc Institute}(2026)]{evo2readme2026}
{Arc Institute}.
\newblock {Evo} 2 {README}: {FP8} and {Transformer Engine} requirements, 2026.
\newblock URL
  \url{https://github.com/ArcInstitute/evo2/blob/53f195997257c56c00e5ef8d33a54f5baad143a6/README.md}.
\newblock Commit 53f1959, 19 June 2026.

\bibitem[Benegas et~al.(2023)Benegas, Batra, and Song]{benegas2023gpn}
Gonzalo Benegas, Sanjit~Singh Batra, and Yun~S. Song.
\newblock {DNA} language models are powerful predictors of genome-wide variant
  effects.
\newblock \emph{Proceedings of the National Academy of Sciences}, 120\penalty0
  (44):\penalty0 e2311219120, 2023.
\newblock \doi{10.1073/pnas.2311219120}.

\bibitem[Nguyen et~al.(2024)Nguyen, Poli, Durrant, Kang, Katrekar, Li, Bartie,
  Thomas, King, Brixi, Sullivan, Ng, Lewis, Lou, Ermon, Baccus,
  Hernandez-Boussard, R{\'e}, Hsu, and Hie]{nguyen2024evo}
Eric Nguyen, Michael Poli, Matthew~G. Durrant, Brian Kang, Dhruva Katrekar,
  David~B. Li, Liam~J. Bartie, Armin~W. Thomas, Samuel~H. King, Garyk Brixi,
  Jeremy Sullivan, Madelena~Y. Ng, Ashley Lewis, Aaron Lou, Stefano Ermon,
  Stephen~A. Baccus, Tina Hernandez-Boussard, Christopher R{\'e}, Patrick~D.
  Hsu, and Brian~L. Hie.
\newblock Sequence modeling and design from molecular to genome scale with
  {Evo}.
\newblock \emph{Science}, 386\penalty0 (6723):\penalty0 eado9336, 2024.
\newblock \doi{10.1126/science.ado9336}.

\bibitem[{Zymrael}(2024)]{vortex}
{Zymrael}.
\newblock {Vortex}: inference and numerics for multi-hybrid {AI} model
  architectures, 2024.
\newblock URL \url{https://github.com/Zymrael/vortex}.
\newblock Apache-2.0; accessed 17 October 2026.

\bibitem[Ku et~al.(2025)Ku, Nguyen, Romero, Brixi, Yang, Vorontsov,
  Taghibakhshi, Lu, Burke, Brockman, Massaroli, R{\'e}, Hsu, Hie, Ermon, and
  Poli]{ku2025stripedhyena2}
Jerome Ku, Eric Nguyen, David~W. Romero, Garyk Brixi, Brandon Yang, Anton
  Vorontsov, Ali Taghibakhshi, Amy~X. Lu, Dave~P. Burke, Greg Brockman, Stefano
  Massaroli, Christopher R{\'e}, Patrick~D. Hsu, Brian~L. Hie, Stefano Ermon,
  and Michael Poli.
\newblock Systems and algorithms for convolutional multi-hybrid language models
  at scale.
\newblock arXiv:2503.01868, 2025.

\bibitem[{NVIDIA}(2025{\natexlab{b}})]{nvidia_transformer_engine}
{NVIDIA}.
\newblock {Transformer Engine}, 2025{\natexlab{b}}.
\newblock URL \url{https://github.com/NVIDIA/TransformerEngine}.
\newblock Version 2.3, released 14 May 2025.

\bibitem[{ongiaf}(2025)]{evo2issue157}
{ongiaf}.
\newblock Evo2\_1b\_base on 5070 get unexpected loss on {Blackwell} ({GitHub}
  issue \#157, {ArcInstitute/evo2}), 2025.
\newblock URL \url{https://github.com/ArcInstitute/evo2/issues/157}.
\newblock Opened 23 July 2025; open as of 17 October 2026.

\bibitem[{dongwei2001}(2025)]{evo2issue132}
{dongwei2001}.
\newblock 7b model passed test while other models all failed ({GitHub} issue
  \#132, {ArcInstitute/evo2}), 2025.
\newblock URL \url{https://github.com/ArcInstitute/evo2/issues/132}.
\newblock Opened 30 May 2025; closed 30 May 2025; last comment 27 August 2025.

\bibitem[{NVIDIA}(2026)]{bionemo_evo2_docs}
{NVIDIA}.
\newblock {BioNeMo} framework: {Evo2} overview (available models, loss and
  {AUC} evaluation), 2026.
\newblock URL
  \url{https://docs.nvidia.com/bionemo-framework/latest/main/developer-guide/bionemo-evo2/bionemo-evo2-Overview/}.
\newblock Undated ``latest'' documentation page; accessed 17 October 2026.

\bibitem[{l1joseph}(2026)]{l1joseph_evo2ft}
{l1joseph}.
\newblock evo2\_ft: {Evo2} ({AMD} {ROCm} port) --- hardware-agnostic inference
  via {FP8}-recipe fine-tuning ({GitHub} repository), 2026.
\newblock URL
  \url{https://github.com/l1joseph/evo2_ft/blob/d026e838a63d1873f9170bc917cc912aed45c248/README.md}.
\newblock README at commit d026e83 (19 September 2026); no preprint; accessed
  19 October 2026.

\bibitem[Ding and Qin(2026)]{ding2026mmaemu}
Xun Ding and Ziyuan Qin.
\newblock {mmaemu}: bit-accurate emulation of {NVIDIA} tensor-core {FP8} {GEMM}
  numerics.
\newblock Zenodo, 2026.
\newblock URL \url{https://zenodo.org/records/23440103}.
\newblock Version 0.1.0, doi:10.5281/zenodo.23440103. Concept DOI (all
  versions): 10.5281/zenodo.23440102.

\bibitem[Xie et~al.(2025)Xie, Xu, Wang, Yang, and Yang]{xie2025mmasim}
Peichen Xie, Shuotao Xu, Yang Wang, Fan Yang, and Mao Yang.
\newblock Bit-accurate modeling of {GPU} matrix multiply-accumulate units:
  Demystifying numerical discrepancy and accuracy.
\newblock arXiv:2511.10909, 2025.
\newblock arXiv v3 (28 Sep 2026); v1 titled ``MMA-Sim: Bit-Accurate Reference
  Model of Tensor Cores and Matrix Cores''. To appear in the 59th IEEE/ACM
  International Symposium on Microarchitecture (MICRO 2026).

\bibitem[Findlay et~al.(2018)Findlay, Daza, Martin, Zhang, Leith, Gasperini,
  Janizek, Huang, Starita, and Shendure]{findlay2018brca1}
Gregory~M. Findlay, Riza~M. Daza, Beth Martin, Melissa~D. Zhang, Anh~P. Leith,
  Molly Gasperini, Joseph~D. Janizek, Xingfan Huang, Lea~M. Starita, and Jay
  Shendure.
\newblock Accurate classification of {BRCA1} variants with saturation genome
  editing.
\newblock \emph{Nature}, 562\penalty0 (7726):\penalty0 217--222, 2018.
\newblock \doi{10.1038/s41586-018-0461-z}.

\bibitem[Ansel et~al.(2024)Ansel, Yang, He, Gimelshein, Jain, Voznesensky, Bao,
  Bell, Berard, Burovski, Chauhan, Chourdia, Constable, Desmaison, DeVito,
  Ellison, Feng, Gong, Gschwind, Hirsh, Huang, Kalambarkar, Kirsch, Lazos,
  Lezcano, Liang, Liang, Lu, Luk, Maher, Pan, Puhrsch, Reso, Saroufim,
  Siraichi, Suk, Zhang, Suo, Tillet, Zhao, Wang, Zhou, Zou, Wang, Mathews, Wen,
  Chanan, Wu, and Chintala]{ansel2024pytorch2}
Jason Ansel, Edward Yang, Horace He, Natalia Gimelshein, Animesh Jain, Michael
  Voznesensky, Bin Bao, Peter Bell, David Berard, Evgeni Burovski, Geeta
  Chauhan, Anjali Chourdia, Will Constable, Alban Desmaison, Zachary DeVito,
  Elias Ellison, Will Feng, Jiong Gong, Michael Gschwind, Brian Hirsh, Sherlock
  Huang, Kshiteej Kalambarkar, Laurent Kirsch, Michael Lazos, Mario Lezcano,
  Yanbo Liang, Jason Liang, Yinghai Lu, C.~K. Luk, Bert Maher, Yunjie Pan,
  Christian Puhrsch, Matthias Reso, Mark Saroufim, Marcos~Yukio Siraichi, Helen
  Suk, Shunting Zhang, Michael Suo, Phil Tillet, Xu~Zhao, Eikan Wang, Keren
  Zhou, Richard Zou, Xiaodong Wang, Ajit Mathews, William Wen, Gregory Chanan,
  Peng Wu, and Soumith Chintala.
\newblock {PyTorch} 2: Faster machine learning through dynamic {Python}
  bytecode transformation and graph compilation.
\newblock In \emph{Proceedings of the 29th ACM International Conference on
  Architectural Support for Programming Languages and Operating Systems, Volume
  2 (ASPLOS '24)}, pages 929--947. ACM, 2024.
\newblock \doi{10.1145/3620665.3640366}.

\bibitem[Blanchard et~al.(2020)Blanchard, Higham, Lopez, Mary, and
  Pranesh]{blanchard2020bfma}
Pierre Blanchard, Nicholas~J. Higham, Florent Lopez, Theo Mary, and Srikara
  Pranesh.
\newblock Mixed precision block fused multiply-add: Error analysis and
  application to {GPU} tensor cores.
\newblock \emph{SIAM Journal on Scientific Computing}, 42\penalty0
  (3):\penalty0 C124--C141, 2020.
\newblock \doi{10.1137/19m1289546}.

\bibitem[Khattak and Mikaitis(2025)]{matlab_tensor_core}
Faizan~A. Khattak and Mantas Mikaitis.
\newblock The {MATLAB Tensor Core}: a set of models of tensor cores written in
  {MATLAB}, 2025.
\newblock URL
  \url{https://github.com/north-numerical-computing/MATLAB-tensor-core}.
\newblock BSD-2-Clause; accessed 17 October 2026.

\bibitem[Fasi et~al.(2021)Fasi, Higham, Mikaitis, and
  Pranesh]{fasi2021numerical}
Massimiliano Fasi, Nicholas~J. Higham, Mantas Mikaitis, and Srikara Pranesh.
\newblock Numerical behavior of {NVIDIA} tensor cores.
\newblock \emph{PeerJ Computer Science}, 7:\penalty0 e330, 2021.
\newblock \doi{10.7717/peerj-cs.330}.

\end{thebibliography}
\end{document}